\documentclass[letterpaper]{article} 
\usepackage{aaai23}  
\usepackage{times}  
\usepackage{helvet}  
\usepackage{courier}  
\usepackage[hyphens]{url}  
\usepackage{graphicx} 
\usepackage{natbib}  
\usepackage{caption} 
\usepackage{algorithm}
\usepackage{algorithmic}
\usepackage[ruled,vlined,algo2e]{algorithm2e}
\usepackage{amsmath}
\usepackage{xcolor} 
\usepackage{newfloat}
\usepackage{listings}
\DeclareCaptionStyle{ruled}{labelfont=normalfont,labelsep=colon,strut=off} 
\floatstyle{ruled}
\newfloat{listing}{tb}{lst}{}
\floatname{listing}{Listing}

\usepackage{booktabs}
\usepackage{scalerel,graphicx,xparse}
\usepackage{algorithm}
\usepackage{algorithmic}
\usepackage{amsmath,amssymb,amsfonts}
\usepackage{soul} 

\title{Metacognitive Reasoning in Energy Based Models using \\ Instance Based Learning Theory} 
\author{
    Tailia Malloy\textsuperscript{\rm 1},
    Prateek Kumar Rajput\textsuperscript{\rm 1}, 
    Serge Lionel Nikiema\textsuperscript{\rm 1} 
     \\ Cleotilde Gonzalez\textsuperscript{\rm 2}, Tegawend\'e F. Bissyand\'e\textsuperscript{\rm 1} 
}
\affiliations{
    \textsuperscript{\rm 1}Interdisciplinary Center for Security, Reliability, and Trust, University of Luxembourg, Luxembourg \\
    \textsuperscript{\rm 2}Department of Social and Decision Sciences, Carnegie Mellon University, Pittsburgh PA 
}

\begin{document}

\maketitle

\begin{abstract}
Metacognition involves reasoning about cognitive processes themselves. An example is in resource allocation where we choose how much time and effort to put into a reasoning task before we begin based on our confidence. Current Artificial Intelligence (AI) systems that rely on Large Language Models (LLMs) cannot estimate their uncertainty about an output without first responding, and cannot dynamically allocate resources to producing an output, making this type of metacognitive process difficult. A recently proposed alternative to classic transformer architectures that addresses these two concerns is the Energy Based Model (EBM) which allows for interpretable uncertainty modeling and dynamic allocation of compute resources. While EBMs can allow for control of these two processes, the actual metacognitive task of determining compute allocation based on uncertainty is not directly addressed. Instance-Based Learning Theory (IBLT) provides an approach to modeling human-like decisions from experience that has previously been applied to predicting human metacognitive reasoning. In this paper we introduce a framework for \textbf{ME}tacognitive \textbf{R}easoning with \textbf{I}nstance-based \textbf{L}earning \textbf{T}heory and \textbf{E}nergy \textbf{D}ynamics (MERITED). Grounded in IBLT, this framework allows for control of the computational effort allocated in an EBM to allow for metacognitive control over reasoning effort based on uncertainty while remaining computationally efficient. This work has two main contributions, the training and open weight sharing of a 191M parameter reasoning EBM, and an implementation of the MERITED framework for dynamic compute allocation using an IBL model.
\end{abstract}

\section{Introduction}
Metacognition, often referred to as `thinking about thinking', can involve several different higher order cognitive processes, often interacting with each other. Despite having direct access to our own cognition, we may struggle to understand metacognitive processes because many of its underlying processes operate outside conscious awareness~\cite{fernandez2000awareness}. Metacognitive processes include estimating uncertainty and allowing that uncertainty to influence how much effort is allocated to complete a task, often without deliberate reflection~\cite{shenhav2017toward}. This limited awareness makes it difficult to determine how AI systems should optimally structure their own metacognitive reasoning. The challenge is compounded by the ability of AI systems to generate sophisticated explanations that appear to describe their metacognitive processes~\cite{barez2025chain}, even when those explanations are not reliably connected to the processes that actually produced their behavior~\cite{turpin2023language}. These post-hoc rationalizations can obscure rather than clarify model behavior~\cite{bilal2025llms,lewis2025analysing}. 

Current methods for reasoning and planning with Large Language Models (LLMs) can involve prompting models to evaluate previously generated outputs, select problem-solving strategies and outline tasks before completing a task~\cite{chen2026towards}. This approach has been related to human metacognition  in a framework that selects reasoning depth based on uncertainty in a way inspired by human cognition~\cite{yang2026think}. However, these approaches operate at the level of model self-prompting during processes such as Chan-of-Thought (CoT) reasoning~\cite{wei2022chain,chen2026towards}. 

Many methods have been developed to explain and understand model output by analyzing these intermediate self-prompting steps \cite{paul2024making, jiang2025safechain}. However, recent work has brought into the question how well these explanations of model behavior can be trusted due to the disconnect between the individual words output by these models during reasoning and the explanations of behavior that are made after reasoning occurs \cite{barez2025chain, arcuschin2025chain}. In this work we focus on the individual token level outputs made by these AI systems by proposing a cognitively inspired metacognitive reasoning method at the level of single outputs.


\section{Background}
In this work we present a framework for \textbf{ME}tacognitive \textbf{R}easoning with \textbf{I}nstance-based \textbf{L}earning \textbf{T}heory and \textbf{E}nergy \textbf{D}ynamics (MERITED), an integration of a cognitive model and an AI system that improves the metacognitive ability of the AI system. The specific AI system is integrated with a cognitive model is called an Energy Based Transformer (EBT), a recent approach that seeks to allow transformer based architectures to better represent their uncertainty and allow for controllable allocation of computational resources~\cite{gladstone2025energy}. While these EBT models are flexible, they do not directly address the metacognitive task of how to optimally allocate computational resources based on experience. To address that, we apply an Instance Based Learning (IBL) model of human learning and decision making onto the task of allocating these resources. Specifically, we use an IBL model to select the optimal `thinking depth' for the EBT model before and during its inference time optimization process. 

A growing body of literature advocates for the integration of complex AI systems like reasoning LLMs, with cognitive models that represent constraints on human information processing~\cite{gonzalez2026toward, gonzalez2025cognitive}. This has been demonstrated in systems that combine cognitive models for predicting human exploratory decision making, with LLMs better predicting exploitative decisions~\cite{nguyen2024predicting}. Other research has incorporated AI models into the processing of complex stimuli, allowing cognitive models to predict human-like behavior without needing human intervention~\cite{malloy2024applying,ajraou2026act}.

However, most of the previous applications of cognitive modeling and AI work to improve the cognitive models rather than the AI systems. The most similar method to what we present in this work is CogRouter introduced by \citealp{yang2026think}, who integrate the ACT-R cognitive architecture~\cite{anderson1993rules,anderson1997act}into the LLM reasoning pipeline to predetermine reasoning depth based on uncertainty estimates. The main difference between this approach and the MERITED framework is that our approach relies on EBMs to perform the uncertainty estimation and computation control, rather than selecting the depth of reasoning. As mentioned in the introduction, the major motivation of this work is to focus on the level of metacognitive reasoning about a single short output representing just a short word or even a single letter. We include a comparison of CogRouter with GPT-2 in the results section demonstrating improved performance on the task we used for evaluation.

The following sections introduce Generative Models, outlining the limitations of some GM methods and how EBMs address these. After this, we provide a theoretical basis for Instance-Based Learning Theory. Finally, we describe the MERITED framework that allows us to instantiate an IBL model of human cognition in resource allocation based on uncertainty for EBMs. 

\section{Generative Models}
\begin{figure*}[htbp]
\centering
\includegraphics[width=0.9\linewidth]{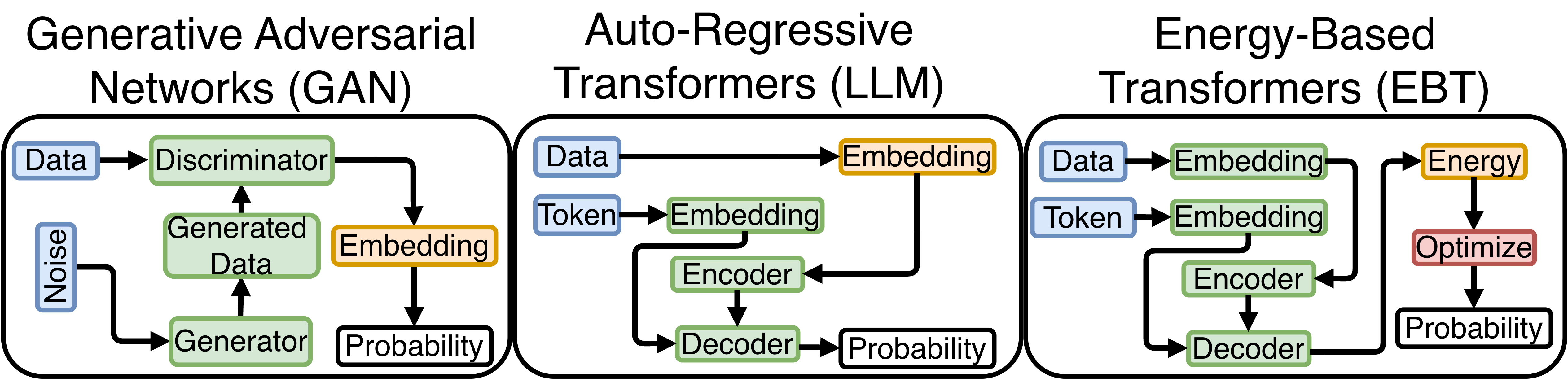}
\caption{Three examples of Generative Models. Previous applications of IBL onto GMs have integrated \textbf{left:} Generative Adversarial Networks (GANs) and \textbf{middle:} Auto-regressive Transformers (ATs). In this work we integrate \textbf{right:} an Energy-Based Model (EBM). Each sub-figure is structured as an input-process-output model, with blue boxes being the input, green boxes being the process, and the orange and black boxes being the output. EBT has an additional step based on the energy function output that involves optimization. The orange box in each sub-figure is the representation that is used to calculate similarity for IBL models of the complex stimuli in a decision making task.}
\label{fig:GenerativeModels}
\end{figure*}
Generative Models (GMs) are a class of machine learning methods that attempt to learn from a dataset by assuming that a probability distribution generated the data and attempting to learn the underlying distribution~\cite{harshvardhan2020comprehensive}. GMs can be categorized into two broad types~\cite{oussidi2018deep}, cost-function based models such as VAEs, and Energy-Based Models (EBMs) where the learned underlying probability distribution is defined as an energy function, such as Bolztman Machines (BMs)~\cite{ackley1985learning}. Figure \ref{fig:GenerativeModels} compares the structure of GANs, Auto-regressive Transformers, and the EBMs that form the basis of the MERITED framework. 

\subsection{Energy Based Models}

EBMs can be viewed as a generalization of classification models, which function similarly under specific assumptions of how models are parameterized~\cite{grathwohl2019your}. Due to the lack of these assumptions, EBMs have historically been difficult to train, but high-dimensional applications have been achieved by leveraging the gradient of the energy function to implicitly select samples from the replay buffer~\cite{du2019implicit}:
\begin{equation}
\tilde{x}^k = \tilde{x}^{k-1} - \gamma/2 \nabla_x E_\theta(\tilde{x}^{k-1}) + \omega^k , \omega^k \sim \mathcal{N}(0,\gamma)
\label{eq:EBM_Sampling}
\end{equation} 
with the process being run through Markov Chain Monte Carlo (MCMC) sampling. 

However, running MCMC sampling until convergence can be computationally expensive, necessitating an approximation method. One common method is Contrastive Divergence, which uses a fixed number of steps for training while still allowing for a variable number of steps to be used at inference time~\cite{song2021train}. This training approach means that we can select different values for the MCMC sample at inference time, thus dynamically controlling the computational resources that are allocated.

\subsection{Energy Based Transformers}
Energy Based Transformers (EBTs) are a recently proposed class of models that combine the transformer architecture with the energy-based learning framework described above~\cite{gladstone2025energy,xu2025energy}. Rather than mapping a context $x$ directly to an output $y$ through a single feed-forward pass, an EBT parameterizes a scalar energy function $E_\theta(x,\hat{y})$ using a transformer backbone, where $\hat{y}$ is a candidate output. The energy $E_\theta(x,\hat{y})$ is trained to be low when the candidate $\hat{y}$ is compatible with the context $x$ and high otherwise, so that a prediction is produced not by direct generation but by iteratively refining a candidate to minimize its energy. 

\begin{minipage}{0.45\textwidth}
\end{minipage}
  \hfill
\begin{minipage}{0.45\textwidth}
  \begin{algorithm}[H] \label{eq:Inference}
    \SetAlFnt{\small}
    \DontPrintSemicolon
    \KwIn{Context $x$, EBT $E_{\theta}(x,\hat{y})$ \\
    	\textbf{Hparams:} Steps $N$, Step Size $\alpha$}
    \vspace{4pt}
      Sample $\hat{y}_{0,j} \sim \mathcal{N}(0, I)$; \\
      \For{$i = 0,\ldots, N-1$} {
        $\hat{y}_{i+1,j} \leftarrow \hat{y}_{i,j} - \alpha \nabla_{\hat{y}_{i,j}} E_{\theta}(x,\hat{y}_{i,j});$ \\
        \lIf{\textit{Condition}}{\textbf{break;}} 
      }
    \Return $\hat{y}^* = \operatorname{argmin}_j E_{\theta}(x,\hat{y}_{N,j});$
    \caption{\scshape Inference}
    \textcolor{red}{\label{alg:Inference}}
  \end{algorithm}
\end{minipage}
\newline 

The Inference algorithm (Algorithm~\ref{alg:Inference}), adapted from ~\citealp{gladstone2025energy}, demonstrates how EBTs allow for dynamic compute allocation. The number of steps $N$ can be set before inference begins and kept static, or if some condition occurs we can break early and stop the sampling. An example of this could be the context $x$ equal to `What is the capital of France' and $Y$ could be [`London', `Paris', `Berlin', `Washington']. The energy of each candidate is randomly initialized $\hat{y}_{0,j} \sim \mathcal{N}(0, I)$ for each $j$ candidate in $Y$. Then we iteratively update this energy prediction based on the previous value $\hat{y}_{i,j}$, and the step size $\alpha$ multiplied by the gradient of the energy function w.r.t the previous value and context $\nabla_{\hat{y}_{i,j}} E_{\theta}(x,\hat{y}_{i,j})$. After doing this same process for each candidate answer, we take the minimum of the vector of energy predictions for each candidate, corresponding to the highest probability based on the estimate. This, therefore, allows for flexibility in the amount of compute resources given to determining any given output relative to the input.

\subsection{Metacognitive Reasoning with EBTs}
Two important features of EBTs make them good targets for metacognitive reasoning and integration with an IBL model: the ability to model uncertainty without needing to produce a response and the dynamic allocation of computational resources during inference. However, these two features also raise new questions: how can uncertainty information be optimally applied to generating outputs? And how should decisions about computational resource allocation be made? In the MERITED framework, both of these questions are addressed using an IBL cognitive model. The following sections detail how an EBT is used to model uncertainty and allocate resources before continuing onto the explanation of the IBL model and ultimately how it is integrated with an EBT within our proposed framework. 

\subsubsection{Modeling Uncertainty}
Making predictions of uncertainty can be difficult for traditional transformer based models, with some approaches relying on iterative prompting~\cite{yadkori2024believe} sampling multiple responses~\cite{zhao2025uncertainty}, asking LLMs to report their uncertainty~\cite{xiong2024can}, or using a different model~\cite{liu2025uncertainty}. However, these approaches typically increase the computational requirements of providing an output, rather than allowing for control over such resources. In EBTs, this issue is addressed by training a model to assign a scalar energy value to inputs that corresponds to their likelihood~\cite{gladstone2025energy}. EBTs model uncertainty by leveraging their probabilistic Energy-Based Models (EBMs) that define a probability distribution using a Boltzmann distribution $p_{\theta}(x) = \exp(-E_\theta(x))/Z(\theta)$ where $Z(\theta) = \int \exp(-E_\theta(x))dx$~\cite{gladstone2025energy, dawid2024introduction, du2019implicit}. 

This allows for a deeper understanding of when the model is more or less sure about its output. However, this probability distribution is not the typical LLM formulation of a probability function over next token outputs, as those probabilities require normalization, and energies are measured in arbitrary units, making direct comparison of different model energy predictions possible~\cite{lecun2006tutorial}. This means that we can directly apply our predictions of model uncertainty to predict the optimal dynamic compute allocation, without needing to first generate an output.

\subsubsection{Dynamic Compute Allocation}
EBTs can control the number of energy optimization steps that they use at inference to find the best solution~\cite{gladstone2025energy}. In real-world problems, there is often a balance between the computational complexity of finding a solution and the need to allocate resources elsewhere, with humans naturally excelling at this type of optimization~\cite{lieder2020resource, malloy2024efficient}. We propose that the dynamic compute allocation feature of EBTs is highly related to metacognitive reasoning. This is because it allows EBTs to think harder about their responses depending on a variety of features of the task at hand. For instance, humans can respond to both low certainty and high impact of decisions by thinking harder about a problem~\cite{gigerenzer2010rationality}: which they can control through metacognition. 

When humans are less confident in their answers or know that their answers are highly impactful, they can allocate more cognitive resources to different areas of a decision making problem~\cite{bates2019adaptive}. This benefit is not only in terms of the runtime efficiency of the model, but also has theoretical benefits in performance by preventing the model from `overthinking'. If the model is sufficiently large and has learned to consistently perform a task by using a high thinking depth, then stopping short isn't expected to improve performance. 

\section{Instance-Based Learning Theory}
Instance-Based Learning Theory (IBLT) provides a set of mathematical constructs that explain the cognitive information processing in humans, allowing for predictions of decision making~\cite{gonzalez2003instance,gonzalez2011instance}. The core component of an IBL model is the memory, consisting of instances $(s,a,x)$ defined by the state $s$, their action $a$ and the outcome $x$. These instance memory components are analogous to more complex and general-purpose cognitive architecture memory components, such as the chunks defined in the Adaptive Control of Thought-Rational (ACT-R)~\cite{anderson1993rules,anderson1997act}. 

IBL groups state-action pairs into options $k=(s,a)$ to define the model memory at time $t$ as $\mathcal{M} = {(k_1,x_1),...,(k_{t-1},x_{t-1})}$. For simplicity, we remove $t$ from the following equations and assume that all values are being calculated at time $t$. IBL models can be used in dynamic decision making tasks by predicting the probability of selecting an option $P_s(k)$ as the soft-max of the predicted utility values $V_{k}$ of each option $\mathcal{A} = (k_1,k_2,...,k_n)$ under consideration~\cite{lebiere2007instance}: 

\begin{equation}
P_s(k) \dfrac{\exp(V(k)/\tau_s)}{\sum_{i\in \mathcal{A}} \exp(V(k_i)/\tau_s)}
\label{eq:ActionSelection}
\end{equation}

where $\tau_s$ is the action selection temperature parameter that sets the degree of stochasticity in action selection. As $\tau$ approaches 0, the model deterministically selects the option with the highest predicted utility value. These option utility values are calculated using the IBL model memory with the \textit{blending} equation~\cite{gonzalez2011instance}: 

\begin{equation}
V(k) = \sum_{i \in \mathcal{M}} P_r(k_i) x_{i}
\label{eq:Blending}
\end{equation}

where $P_r(k_i)$ is the probability of retrieving the instance $i$ from memory while predicting the utility value of option $k$, and $x_i$ is the utility of option $k_i$ in memory. The probability $P_r(k)$ of retrieving an instance $k$ from memory while evaluating the utility value of an option is defined as a softmax of the instance activations $A(k)$ 

\begin{equation}
A(k) = ln(D) +  \sum_{f \in F} \alpha_f S_f(k^f,k_i^f)  + \sigma \ln \dfrac{1 - \xi}{\xi}
\label{eq:Activation}
\end{equation}
where $D$ is the decay constant that is defined based on the decay parameter $\delta$ as $D = \sum_{t'\in T}(t-t')^{-d}$, $\sigma$ is the noise parameter. The similarity function $S(k,k_i)$ is used to compare the similarity between the option under consideration $k$ and options in memory $k_i$. Default parameter values are set to $(\sigma=0.5, \delta=0.25, \tau_r= 0.5,\tau=\sigma \sqrt{2})$ based on previous studies of human behavior with IBL~\cite{gonzalez2011instance}.

\section{EBT-IBL Integration}
Our proposed MERITED framework is described as an integration of an IBL cognitive model and the EBT model. Previous integrations of IBL models with GMs have focused on cost-function-based models to provide similarity measures and representations of complex stimuli~\cite{malloy2024applying}. These applications tend to focus on using GMs to expand the applicability of cognitive models into domains with more complex stimuli. In this work, we propose an integration of IBL with an EBT that not only serves to provide a stimuli representation and similarity measure but also allows the IBL model to expand the metacognitive reasoning abilities of the EBT. Thus, our application is to improve the performance of the AI system through an integration of a metacognitive reasoning ability afforded by the IBL model. 

The central property of EBTs that we exploit for our IBL integration is that the learned energy function $E_\theta(x,\hat{y})$ provides a natural task-grounded feature space for similarity calculations. Additionally, the task domain of question answering provides a more concrete definition of model accuracy, allowing us to directly compare the task performance of models and provide an IBL model with a utility history of depth selections. For a given context $x$, the energy landscape induced by the optimization trajectory $\{\hat{y}_0, \hat{y}_1, \dots, \hat{y}_N\}$ characterizes how difficult that input is to solve: inputs whose energy decreases rapidly and settles to a low, flat minimum are ``easy'' and require little compute, whereas inputs with slowly decreasing, high, or non-convex energy profiles are ``hard'' and benefit from additional MCMC optimization steps. 

\begin{figure*}[htbp]
\centering
\includegraphics[width=0.85\linewidth]{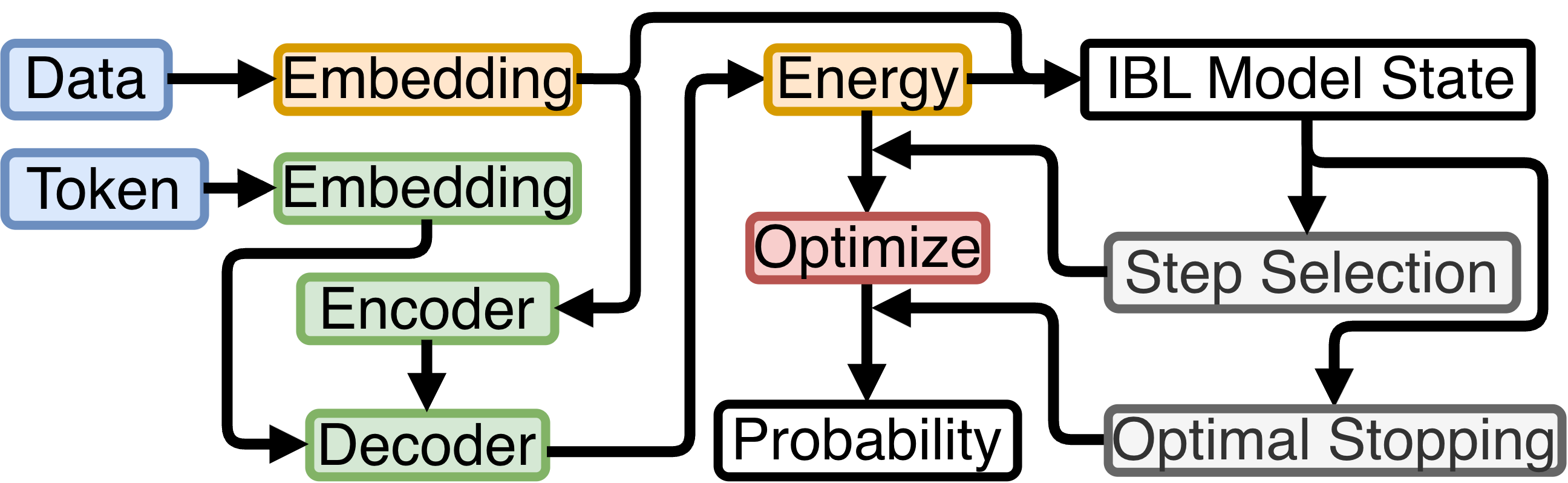}
\caption{Orange boxes represent the embedding and energy distribution used by the IBL model. Here, the decision made by the IBL model is either what depth to use (step selection) or whether to stop optimization (optimal stopping). The EBT generates the energy distribution and question embeddings fed into IBL. The final framework output from processing the data and candidate tokens after the optimization step is a probability distribution.} 
\label{fig:MERITED-Diagram}
\end{figure*}

Figure \ref{fig:MERITED-Diagram} shows a diagram of the MERITED framework. The only change to the input-process-output structure from an EBT without the IBL integration is that the data embedding and energy function define the IBL model state, which then determines the optimization strategy by either selecting it from different options or deciding when to stop optimization. 
These are the two different ways of integrating the IBL model predictions into MERITED. Firstly, we can select the MCMC step size based on the initial energy distribution and question embeddings. This mirrors the method of statically allocating reasoning depth, but this is done before reasoning begins based on the features of the problem. The alternative to this is to evaluate whether MCMC optimization should continue at each step, allowing for more fine-tuned control over reasoning progression but requiring more computation.. 

\subsection{EBT Similarity Measure}
At each timestep of the IBL model experience, the EBT takes as input the context $x$ and candidate answers $y=(y_1,y_2,...,y_n)$ to produce the energy distribution $E_{\theta}(x,\hat{y}_{N,j})$ and the input embedding $E_m(x)=\mathbb{R}^d$. These two vectors make up the IBL model state, which is combined with a set of MCMC step size selections $A=(\alpha_1,\alpha_2,...,\alpha_m)$ in the case of step selection and stop/continue $A=(\alpha_S, \alpha_C)$ in the case of optimal stopping. These actions produce the options $(k_1,k_2,...,k_m)$ allowing for the calculation of similarity between instances in memory as: 
\begin{equation}
\begin{split}
& S_{\mathbb{E}_{\theta}}(E_{\theta}(k_x,k_{\hat{y}}),E_{\theta}(k_{i,x},k_{i,\hat{y}}) + S_C(E(k_x),E(k_{i,x})) + \\& S_C(E(k_y),E(k_{i,y})) + S_I(k_{\alpha},k_{i,\alpha}) = S(k,k_i)
\end{split}
\label{eq:EBM_Similarity}
\end{equation} 
where $S_{\mathbb{E}_{\theta}}$ is the euclidean distance $S_C$ is the cosine similarity, and $S_I$ is the indicator function. 

\subsection{MCMC Step Selection}
When the IBL model is used to select the number of MCMC steps and the step size, it evaluates the options according to: 

$$N,\alpha \leftarrow P_s(k), \text{ w.r.t } S_{E_{\theta}(x,\hat{y})}(k,k'),  k = (N,\alpha,x)$$
where $S_{E_{\theta}(x,\hat{y})}(k,k')$ is the similarity function defined by the energy function output of the EBT model parameterized by $\theta$, with options $k = (N,\alpha,x)$ representing pairs of contexts $x$, step sizes $\alpha$, and number of steps $N$.

Because the energy function supplies both the representation and the similarity metric, the IBL model can predict metacognitive control decisions for the EBT purely from experience with similar past inputs. We investigate two such uses. In the first (Algorithm~\ref{alg:StepSelection}), the IBL model treats the number of MCMC steps $N$ and the step size $\alpha$ as part of the option $k=(N,\alpha,x)$ and selects them jointly via the option-selection probability $P_s(k)$ of Eq.~\ref{eq:ActionSelection}, blending over past experiences weighted by the energy-based similarity $S_{E_\theta(x,\hat{y})}(k,k')$; this predicts \emph{up front} how much compute a new input should receive. 

\begin{algorithm}[H] \label{alg:LLMITL}
    \SetAlFnt{\small}
    \DontPrintSemicolon
    \KwIn{Context $x$, EBT $E_{\theta}(x,\hat{y})$ IBL Model $P_s(k)$, \\ Similarity Metric $S_{E_{\theta}(x,\hat{y})}(k,k')$\\
    	\textbf{Hparams:} Step Increases $n$, Thresholds $t$}
    \vspace{4pt}
    $N,\alpha \leftarrow P_s(k), \text{ w.r.t } S_{E_{\theta}(x,\hat{y})}(k,k'),  k = (N,\alpha,x)$\\
      Sample $\hat{y}_{0,j} \sim \mathcal{N}(0, I)$; \\
      \For{$i = 0,\ldots, N-1$} {
        $\hat{y}_{i+1,j} \leftarrow \hat{y}_{i,j} - \alpha \nabla_{\hat{y}_{i,j}} E_{\theta}(x,\hat{y}_{i,j});$
      }
    \Return $\hat{y}^* = \operatorname{argmin}_j E_{\theta}(x,\hat{y}_{N,j});$ \\
    \uIf{$M(y) < t_0$}{
        $N \leftarrow N + n_0 $\\
    }
    \uElseIf{$M(\hat{y}^*) < t_1$}{
        $N \leftarrow N + n_1 $\\
    }
    \caption{\scshape IBL Step and Step Size Selection for EBT MCMC}
    \textcolor{red}{\label{alg:StepSelection}}
\end{algorithm}

\subsection{MCMC Optimal Stopping}
The other alternative we investigate is to use the IBL model as an optimal-stopping decision maker, where it is queried at each timestep of the MCMC to evaluate whether the process can stop. This can be formulated as an optimal stopping problem, to which IBL models have previously been applied~\cite{bugbee2022making}. Since there are no utility signals in optimal stopping tasks, the IBL model is adjusted to assign credit for the final stop decision to previous instances where the decision was to continue, as is done in other environments with delayed rewards~\cite{nguyen2023credit}.

\begin{algorithm}[H]
    \SetAlFnt{\small}
    \DontPrintSemicolon
    \KwIn{Context $x$, EBT $E_{\theta}(x,\hat{y})$ IBL Model $P_s(k)$, \\ Similarity Metric $S_{E_{\theta}(x,\hat{y})}(k,k')$\\
\textbf{Hparams:} Max Steps $N$, Step Size $\alpha$, Stop Threshold $\tau$}    \vspace{4pt}
    Sample $\hat{y}_{0,j} \sim \mathcal{N}(0, I)$; \\
    \For{$i = 0,\ldots, N-1$} {
    $\hat{y}_{i+1,j} \leftarrow \hat{y}_{i,j} - \alpha \nabla_{\hat{y}_{i,j}} E_{\theta}(x,\hat{y}_{i,j});$\\
    $\Delta_{i,j} \leftarrow E_{\theta}(x,\hat{y}_{i,j}) - E_{\theta}(x,\hat{y}_{i+1,j});$\\
    $c \leftarrow P_s(k_i), \text{ w.r.t } S_{E_{\theta}(x,\hat{y})}(k_i,k'), \\ k_i = (i,\alpha,x,\hat{y}_{i+1,j});$\\
    \lIf{$c < \tau$}{\textbf{break}}
    }
    \Return $\hat{y}^* = \operatorname{argmin}_j E_{\theta}(x,\hat{y}_{i+1,j});$

    \caption{\scshape IBL Optimal Stopping for EBT MCMC}
    \textcolor{red}{\label{alg:OptimalStopping}}
\end{algorithm}

In the second (Algorithm~\ref{alg:OptimalStopping}), the IBL model is queried at each MCMC step in an optimal-stopping fashion, using the same similarity metric to decide whether the marginal expected reduction in energy justifies continuing the minimization or whether the current candidate should be accepted and the process stopped. The first strategy amortizes the decision into a single prediction, while the second adapts the compute budget online as the energy trajectory unfolds; both rely on the EBT energy function as the common substrate for measuring similarity between the current input and prior experience.

\section{Experimentation}
We are interested in evaluating whether the MERITED framework can allocate EBT inference depth to produce higher performance in question answering tasks compared to statically setting it to any one depth. Ideally, this IBL selection method of the MERITED framework would also reduce the runtime of the model compared to high static depth settings. We compare five static EBT depths for MCMC alongside the two IBL methods that we propose, optimal stopping and depth selection. As a baseline, we also use a similarly structured LLM with a similar number of parameters and training tokens, namely GPT-2. All IBL model parameters are set to their defaults $(\sigma=0.5, \delta=0.25, \tau_r= 0.5,\tau=\sigma \sqrt{2})$. and the EBT models function identically in the MERITED framework comparison as they do in the static depth setting. 

\subsection{Tasks and Data Splits}
We use two elementary science question and answering tasks, both available online\footnote{\url{https://huggingface.co/datasets/allenai}}. The first of these is simpler, the ARC-Easy Dataset, also used for model training, which contains 570 examples within the validation split (567 questions have four choices, one has three, and two have five). The second task is more challenging, the OpenBookQA Dataset, which uses 4,957 four-choice example questions from the main training split. Thus, these OpenBookQA results characterize the experimental policy instead of a test-set generalization estimate. For OpenBookQA, the question prompt includes the labeled answer choices and asks for one label from A to D as the solution to the question. Compared to ARC-EASY, these questions require more multi-step reasoning in general. 

For evaluation, we generate splits of 100 questions for the ARC-Easy dataset and 500 questions for the OpenBookQA dataset. There are 10 different random seed runs for each evaluation. The same questions are used for the evaluation of all models. Different random seeds are applied to each evaluation run. To provide the required initial memory for the IBL model depth selection, we use a sample of 20 questions with corresponding optimal depth selections as the pre-populated memory of the IBL model. 

\subsection{Models, Inference, and Scoring}
The EBT condition uses the 191M-parameter model configuration, and the GPT-NeoX-20B tokenizer, see Table \ref{tab:ebt-gpt2-config} of the supplementary materials for additional training details. Static EBT inference is evaluated at MCMC depths $\mathcal{D}=\{1,2,4,8,10\}$. The baseline is the pretrained Hugging Face \texttt{gpt2} checkpoint (GPT-2 small, approx 124M parameters). All conditions for evaluation run on the PyTorch MPS backend. EBT randomness is disabled during scoring, and item/choice seeds are derived deterministically from the experiment seed.

\begin{figure*}[ht!]
\centering
\includegraphics[width=\linewidth]{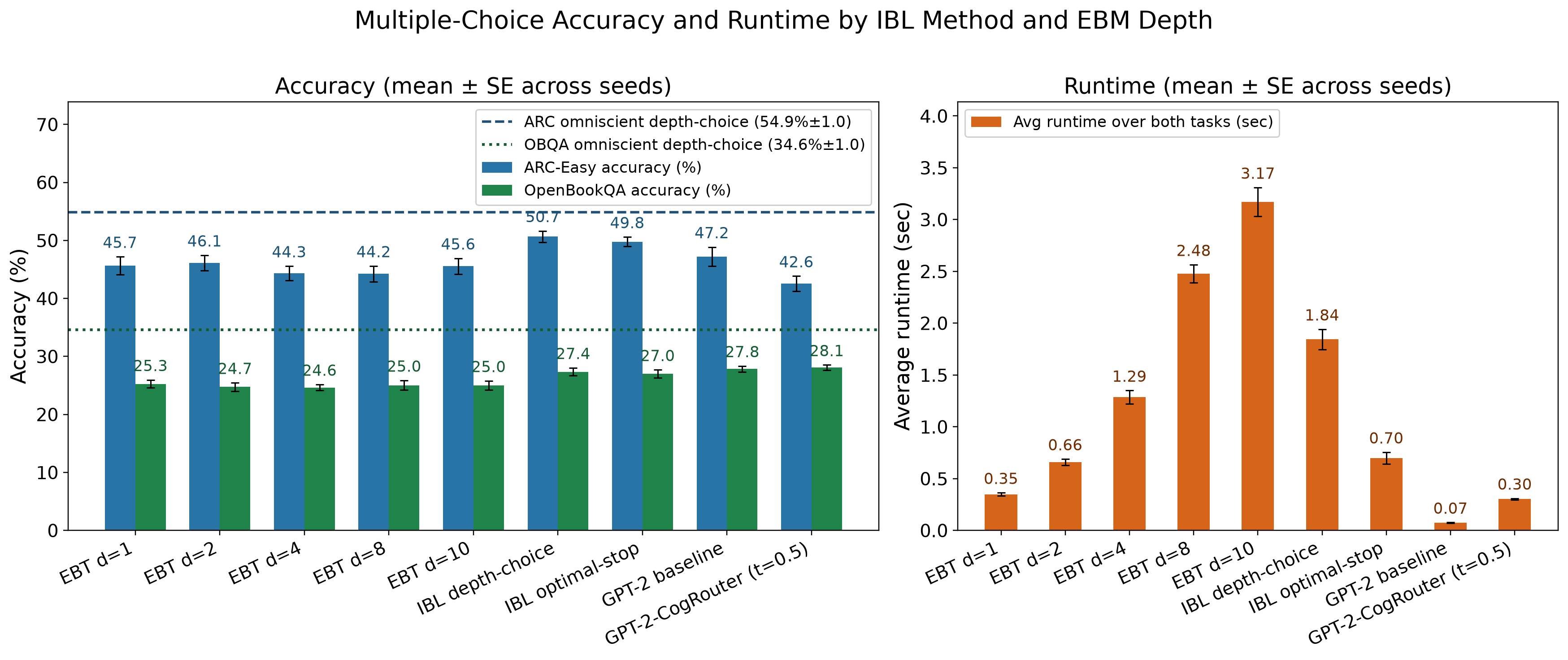}
\caption{Left Panel: Question response accuracy for the ARC-Easy (blue) and OpenBookQA (green) datasets. Right Panel: Orange bars indicate the average runtime for answering the question across both tasks in seconds. The dashed line represents the highest accuracy by selecting from all EBT depths for the ARC dataset, and the dotted line for the OpenBookQA dataset. Random choice performance for all questions is 25\%. }
\label{fig:performance}
\end{figure*}

For both models, a question is formatted as "Question: $q$ Answer:", and each candidate is appended separately. At depth $d$, the score for candidate $y$ is the sum of its conditional token log-probabilities,
\begin{equation}
L_d(y\mid q)=\sum_{j=1}^{|y|}\log p_d(y_j\mid q,y_{<j}),
\end{equation}
and the prediction is $\arg\max_y L_d(y\mid q)$. The primary accuracy shown in Figure~\ref{fig:performance} and Table~\ref{tab:combined_two_task_best_sweep} uses this un-normalized sum. Length-normalized scores are also recorded, but they are used only as a secondary hyperparameter tie-break and are not the plotted metric.

\subsection{IBL State, Memory, and Depth Control}

For question $q$ at depth $d$, the IBL state is
\begin{equation}
s_d(q)=\left(\mathbf{p}_d(q),\mathbf{e}(q)\right),
\end{equation}
where $\mathbf{p}_d(q)=\operatorname{softmax}(-\mathbf{E}_d(q))$ is the distribution formed from the mean EBT answer-token energy for each candidate and $\mathbf{e}(q)$ is the $\ell_2$-normalized mean of the EBT token embeddings for the question prompt. For ARC-Easy this prompt representation contains only the question; for OpenBookQA it also contains the rendered choices and label-only response instruction. The following weighted partial similarities are used for the IBL instance features:
\begin{equation}
S(s,s')=\lambda_E\left(1-\tfrac{1}{2}\lVert\mathbf{p}-\mathbf{p}'\rVert_1\right)
+(1-\lambda_E)\left(\tfrac{1+\cos(\mathbf{e},\mathbf{e}')}{2}\right),
\label{eq:experiment-state-similarity}
\end{equation}
with the energy weight $\lambda_E$ selected by the sweep. MERITED is instantiated alongside the PyIBL python library\footnote{\url{https://pypi.org/project/pyibl/}} which incorporates these weighted partial similarities into instance activation under the swept mismatch penalty.

The \emph{depth-choice} policy treats each $d\in\mathcal{D}$ as an action. Each of the 20 memory questions is evaluated at all five depths, producing 100 state-action-outcome instances. The state supplied for every action contains that memory question's depth-1 energy distribution, and the outcome is $+1$ when the EBT answer at that depth is correct and $-1$ otherwise. At test time, MERITED first computes the depth-1 state, queries IBL once, and runs the selected depth as a separate trajectory when $d>1$.

The \emph{optimal-stop} policy uses a separate two-action IBL agent at every completed depth $d\in\{1,\ldots,9\}$. For each memory question and decision depth, the \textsc{stop} outcome is $+1$ or $-1$ according to correctness at $d$, while the \textsc{continue} outcome is determined by correctness at $d+1$. This yields $20\times 2\times 9=360$ memory instances. During evaluation, the policy observes the current energy distribution and chooses \textsc{stop} or \textsc{continue} after each MCMC step, with a forced stop after depth 10. All memory instances are inserted at model time zero, and evaluation uses history-less choices so test questions do not update memory.

\subsection{Metrics, Timing, and Statistical Analysis}
We report primary multiple-choice accuracy and mean wall-clock inference time per question. Static EBT and GPT-2 times include candidate scoring. Depth-choice time includes depth-1 state acquisition, the IBL query, and execution at the selected depth. For the archived optimal-stop implementation, the current-depth energy distributions supplied to IBL are read from the test trajectories cached before timing; therefore, its time includes dynamically terminated candidate scoring and all intervening IBL queries, but not the generation of those state distributions. PyTorch is synchronized before and after timing. Model loading, question-embedding construction, IBL memory construction, hyperparameter search, and plotting are also excluded. 

The omniscient depth-choice oracle marks an item correct if any $d\in\mathcal{D}$ answers it correctly; it is an upper bound that uses the gold answer to select depth after inference and is therefore not deployable. The combined test pools correct counts across tasks, whereas mean accuracy is the unweighted mean of the two task accuracies. No multiple-comparison correction is applied, and the tests do not account for selecting each IBL policy from 450 configurations on the same evaluation data; the $p$-values should therefore be interpreted descriptively. Because there is one split and one seed, the reported standard deviations and standard errors are descriptive variation across the two task-level values, not uncertainty across repeated runs.

\subsection{Results}
All code required for model training and testing, as well as model weights and scripts for generating results figures, are made available online \footnote{\url{https://github.com/TailiaReganMalloy/instance-energy-based-learning}}. Figure~\ref{fig:performance} and Table~\ref{tab:combined_two_task_best_sweep} show that both MERITED policies improve over every static-depth setting in unweighted mean accuracy. IBL depth choice obtains 51.8\%, compared with 52.2\% for IBL optimal stopping, 48.2\% for the best static means (depths 2 and 10), and 44.0\% for GPT-2. Under the implemented one-sided test, the pooled improvement of depth choice over GPT-2 has $p=0.036$, while the pooled optimal-stop comparison has $p=0.068$; these values are descriptive given the selection caveat above.

On ARC-Easy, both MERITED policies reach 58\% accuracy. This is 3 percentage points above the best static EBT result (55\% at depths 2 and 10), 16 points above GPT-2 (42\%), and 2 points below the 60\% omniscient depth-choice oracle. Both MERITED comparisons with GPT-2 have $p=0.012$. The selected depth-choice policy uses mean depth 5.62, while optimal stopping terminates at mean depth 2.69. Thus, the online policy attains the same ARC-Easy accuracy as one-shot depth allocation while using substantially fewer MCMC steps.

On OpenBookQA, depth choice is the strongest observed condition at 29.4\%, followed by optimal stopping at 28.4\%, static depth 8 at 27.2\%, and GPT-2 at 26.8\%. The gains over GPT-2 are not significant ($p=0.180$ and $p=0.286$, respectively). The corresponding mean selected depths are 6.22 for depth choice and 1.95 for optimal stopping. The 36.4\% depth-choice oracle is 7 points above the learned depth-choice policy, showing that different static depths solve complementary items but that the 20-question IBL memory does not reliably identify all of them. When every integer depth from 1 to 10 is available, the optimal-stop oracle is 40.8\%; this unplotted ceiling also shows that most OpenBookQA items remain incorrect at every tested depth.

The reported timing results in the left side of Figure \ref{fig:performance} distinguish the two control strategies. Mean per-question runtime is 2.0883 seconds for depth choice and 0.73 seconds for optimal stopping. Within the cached-state timing boundary, optimal stopping is 64.9\% faster than depth choice and 76.1\% faster than static depth 10 (3.06 seconds), while losing only 0.50 points of unweighted mean accuracy relative to depth choice. Depth choice is 32.0\% faster than depth 10 and improves mean accuracy by 2.9 points. Thus, one-shot depth choice gives the highest observed accuracy, while optimal stopping is the faster measured policy.

To compare against an alternative approach for controlling model reasoning, we evaluated GPT-2-CogRouter \cite{yang2026think} on the same evaluation questions used for the static EBT depths, the two MERITED policies, and the GPT-2 baseline. GPT-2-CogRouter first scores each candidate answer directly; if the margin between the top two candidates falls below a fixed confidence threshold, the question is re-scored using a Level 2 situational-awareness prompt adapted from the original CogRouter framework, and the higher-confidence answer is kept otherwise. The exact configuration and confidence threshold used for this comparison are reported in the supplementary materials in Table~\ref{tab:cogrouter-config}. Additionally, we include all implementation alterations in our online codebase\footnote{\url{https://github.com/TailiaReganMalloy/instance-energy-based-learning}}. 

Comparing the alternative method CogRouter, both of the MERITED models outperform this approach though to a lesser degree than the standard GPT-2 model. It should be noted that this method was not designed for the type of question answering response where a single next token is evaluated for accuracy. Rather, this approach is intended to control the thinking depth of the model. Because of this it was necessary to adjust the implementation from the original work to function within the parameters of this comparison. Additionally, because the GPT-2 model is small and unlikely to display the type of improvements from internal reasoning that larger models would, it is not clear from these results if the same pattern would hold on larger models. 

\begin{table}[t]
\centering
\small
\begin{tabular}{lrrrr}
\hline
Model & ARC (\%) & OB (\%) &Std & Time (s) \\
\hline
EBT d=1 & 52.00 & 26.40  & 18.10 & 0.35  \\
EBT d=2 & 55.00$^{*}$ & 26.60  & 20.08 & 0.65  \\
EBT d=4 & 53.00 & 25.60 & 19.37 & 1.25 \\
EBT d=8 & 51.00 & 27.20  & 16.83 & 2.45  \\
EBT d=10 & 55.00$^{*}$ & 26.60  & 20.08 & 3.07 \\
MERITED Depth & \textbf{58.00}$^{*}$ & \textbf{29.40}  & 20.22 & 2.09  \\
MERITED Stop & \textbf{58.00}$^{*}$ & 28.40 & 20.93 & 0.73 \\
GPT-2 & 42.00 & 26.80  & 10.75 & \textbf{0.07} \\
GPT-2-CogRouter & 45.00 & 23.90  & 10.75 & \textbf{0.07} \\
\hline
Omniscient & 60.00 & 36.40 & -- &  -- \\
\hline
\end{tabular}
\caption{ARC accuracy, OB accuracy, Accuracy standard deviation, and runtime for each of the 5 static MCMC depth models, the two MERITED methods, and the GPT-2 baseline performance. Omniscient performance is theoretical limit of MERITED performance.}
\label{tab:combined_two_task_best_sweep}
\end{table}

\section{Conclusion}
We introduce MERITED, a framework for Metacognitive Reasoning using Instance-Based Learning Theory and Energy Dynamics. By integrating an Instance-Based Learning (IBL) model with an Energy-Based Transformer (EBT), MERITED addresses a central challenge for EBTs: determining how much computational resources to allocate at inference time. MERITED treats this problem as a metacognitive task with two components: predicting task difficulty before attempting a problem and dynamically allocating computational resources based on that prediction. We examine two applications of IBL to this task: selecting the initial amount of cognitive resources and deciding when to finish computation through optimal stopping. We evaluated MERITED using a newly trained 191 million parameter EBT and compared its performance with static computation allocation strategies and a similarly sized transformer-based model (GPT-2). Across two question-answering tasks, MERITED improved performance compared to static allocation, while reducing response time compared to a strategy that allocated a high level of computation to every question. 

These results demonstrate how cognitive models can improve the operation of AI systems by providing mechanisms for adaptive metacognitive control. The motivation for this type of integration is related to the goal of complementarity in human-AI interaction research, which seeks to allocate functions according to the relative strengths of humans and AI systems~\cite{gonzalez2025cognitive}. MERITED applies a similar principle within an integrated computational system, where the IBL model performs metacognitive control, while the EBT performs question answering using computational resources allocated by the IBL model. This direction is novel compared to many previous integrations of cognitive models and generative models, that use representations learned by generative models to extend the representations and similarity functions involving complex stimuli~\cite{gonzalez2026toward}. MERITED reverses this approach by using a cognitive model to improve the adaptive functioning of an AI system. 

Several limitations qualify these findings. First, the number of parameters in the EBT is 191 million, substantially fewer than current transformer based LLMs with 10s of billions of parameters. It remains unclear whether it is possible to generalize to larger model sizes and see the same improvement in question answering ability afforded by the IBL selection of optimal stopping of MCMC steps or the initial depth choice. Instead, these results seek to demonstrate a proof of concept for a smaller model by comparing it to a similarly sized model that is also trained on roughly the same dataset size (GPT-2). Second, while we do use two different question answering tasks with different difficulties, the models themselves are not using higher level thinking patters like chain-of-thought or graph-of-thought, and comparing metacognitive control over this type of reasoning remains unexplored in this work. Future research should evaluate MERITED with larger EBTs with billions of parameters, comparing performance in direct answering tasks with more reasoning based methods, and exploring more complex task domains such as agentic AI, logical and mathematical reasoning, and computer programming. 

\clearpage

\section*{Acknowledgements}
This research was supported by Google.org through the provision of Google Cloud Platform credits used to conduct our experiments.

\bibliography{aaai23}
\section*{Supplementary Materials}
In the original paper, there was an additional optionall sampling step that meant each loop of the MCMC sampling could be run multiple times. This was related to the `answer confirmation' motivation of EBTs, which we do not focus on in this work.

\begin{algorithm}[H] \label{eq:InferenceOriginal}
\SetAlFnt{\small}
\DontPrintSemicolon
\KwIn{Context $x$, EBT $E_{\theta}(x,\hat{y})$ \\
    \textbf{Hparams:} Steps $N$, Step Size $\alpha$, Samples $M$}
\vspace{4pt}
\For{$j = 1,\ldots, M$} {\nobreak
  Sample $\hat{y}_{0,j} \sim \mathcal{N}(0, I)$; \\
  \For{$i = 0,\ldots, N-1$} {
    $\hat{y}_{i+1,j} \leftarrow \hat{y}_{i,j} - \alpha \nabla_{\hat{y}_{i,j}} E_{\theta}(x,\hat{y}_{i,j});$
  }
}
\Return $\hat{y}^* = \operatorname{argmin}_j E_{\theta}(x,\hat{y}_{N,j});$
\caption{\scshape Inference}
\textcolor{red}{\label{alg:InferenceOriginal}}
\end{algorithm}

\begin{algorithm}[H] \label{eq:TrainingOriginal}
\SetAlFnt{\small}
\DontPrintSemicolon
\KwIn{Context $x$, Target $y$, EBT $E_{\theta}(x,\hat{y})$ \\
    \textbf{Hparams:} Steps $N$, Step Size $\alpha$, Loss $J(\cdot)$}
\vspace{4pt}
\For{$j = 1,\ldots, M$} {\nobreak
Sample $\hat{y}_0 \sim \mathcal{N}(0, I)$; \\
\For{$i = 0,\ldots, N-1$} {
  $\hat{y}_{i+1} \leftarrow \hat{y}_i - \alpha \nabla_{\hat{y}_i} E_{\theta}(x,\hat{y}_i);$
}
}
$\mathcal{L} \leftarrow J(\hat{y}_N, y);$ \\
\Return $\mathcal{L}$, update $E_{\theta}$;
\caption{\scshape Training}
\textcolor{red}{\label{alg:TrainingOriginal}}
\end{algorithm}

Algorithm~\ref{alg:TrainingOriginal} details how EBTs are trained to predict the output $y$ given the context $x$ by learning to minimize the energy $E_\theta(x,y)$.

\begin{table}[H]
\centering
\setlength{\tabcolsep}{3pt}
\caption{GPT-2-CogRouter evaluation configuration. This table describes the router wrapper used in the comparison, not a separately fine-tuned GPT-2 checkpoint.}
\label{tab:cogrouter-config}
\begin{tabular}{@{}lp{0.68\columnwidth}@{}}
\toprule
Component & Configuration \\
\midrule
Base model & Same pretrained GPT-2 small checkpoint as Table 3 \\
Level 1 & Fast direct GPT-2 candidate scoring \\
Routing signal & Difference between the two largest softmax-normalized candidate scores \\
Confidence threshold & 0.50 margin; lower-confidence questions are routed for additional reasoning \\
Routed reasoning & Level 2 situational-awareness prompt containing the question and candidate choices \\
Final answer & Candidate with the highest rescored GPT-2 log-probability \\
Runtime & Includes the initial scoring pass and any routed rescoring pass \\
\bottomrule
\end{tabular}
\end{table}

\onecolumn 
\begin{table}[t]
\centering
\caption{Model architecture and training configuration for the 191M Energy-Based
Transformer (EBT) and the GPT-2 small baseline (SLM) used in the ARC-Easy
multiple-choice comparison. Dashes (---) denote parameters that do not apply to the standard
autoregressive baseline.}
\label{tab:ebt-gpt2-config}
\begin{tabular}{lcc}
\toprule
 & \textbf{EBT-191M} & \textbf{GPT-2 small (SLM)} \\
\midrule
\multicolumn{3}{l}{\emph{Architecture}} \\
Parameters                 & $\sim$191M                    & 124M \\
Transformer blocks         & 12                            & 12 \\
Embedding dimension        & 1024                          & 768 \\
Attention heads            & 16                            & 12 \\
Head dimension             & 64                            & 64 \\
Context / sequence length  & 512                           & 1024 \\
Vocabulary size            & 50{,}277                      & 50{,}257 \\
Tokenizer                  & GPT-NeoX-20B                  & GPT-2 BPE \\
\midrule
\multicolumn{3}{l}{\emph{Energy-based inference (MCMC)}} \\
MCMC steps (training)      & 2                             & --- \\
MCMC step size $\alpha$    & 0.25                          & --- \\
Langevin dynamics noise    & 1.0                           & --- \\
Step-size scaling          & energy-scaled ($\tau{=}1.0$)  & --- \\
Eval ``thinking'' depths   & $\{1,2,4,8,10\}$                   & --- \\
\midrule
\multicolumn{3}{l}{\emph{Pretraining}} \\
Training tokens            & 3.0B (target; $\geq$2.5B min.)& WebText (OpenAI) \\
Dataset                    & ClimbMix                      & WebText \\
Precision                  & bf16                          & --- \\
Peak learning rate         & $3\times10^{-4}$              & --- \\
Min.\ LR fraction          & 0.1                           & --- \\
Warmup steps               & 300                           & --- \\
Weight decay               & 0.1                           & --- \\
Gradient clip              & 0.5                           & --- \\
Global batch (sequences)   & 512 ($8\times8\times8$)       & --- \\
Hardware                   & $8\times$B200                 & --- \\
Checkpoint used            & pretrain-base, step 7000      & pretrained \\
\bottomrule
\end{tabular}
\end{table}

\begin{table}[t]
\centering
\small
\begin{tabular}{lrrrrrr}
\hline
Model & ARC Acc (\%) & OB Acc (\%) & Mean Acc (\%) & Acc Std & Avg Runtime (s) & Runtime Std \\
\hline
EBT d=1 & 52.00 & 26.40 & 39.20 & 18.10 & 0.35 & 0.02 \\
EBT d=2 & 55.00$^{*}$ & 26.60 & 40.80 & 20.08 & 0.65 & 0.07 \\
EBT d=4 & 53.00 & 25.60 & 39.30 & 19.37 & 1.25 & 0.15 \\
EBT d=8 & 51.00 & 27.20 & 39.10 & 16.83 & 2.45 & 0.29 \\
EBT d=10 & 55.00$^{*}$ & 26.60 & 40.80 & 20.08 & 3.07 & 0.39 \\
MERITED depth-choice & \textbf{58.00}$^{*}$ & \textbf{29.40} & \textbf{43.70$^{*}$} & 20.22 & 2.09 & 0.35 \\
MERITED optimal-stop & \textbf{58.00}$^{*}$ & 28.40 & 43.20 & 20.93 & 0.73 & 0.07 \\
GPT-2 baseline & 42.00 & 26.80 & 34.40 & 10.75 & \textbf{0.07} & 0.00 \\
GPT-2-CogRouter baseline & 44.00 & 28.20 & 34.40 & 10.75 & \textbf{0.07} & 0.00 \\
\hline
Omniscient depth-choice & 60.00 & 36.40 & -- & -- & -- & -- \\
\hline
\end{tabular}
\caption{Combined two-task best-sweep comparison across ARC-Easy and OBQA. GPT-2 is the baseline row. Asterisk marks one-sided significance over GPT-2 baseline at $\alpha=0.05$. Acc Std and Runtime Std are computed across the two task-level values for each model. Bolding represents best performance within column, ignoring the hypothetical optimal depth-choice.}
\label{tab:combined_two_task_best_sweep_full}
\end{table}

\end{document}